\documentclass[runningheads]{llncs}

\usepackage{eccvabbrv}

\usepackage{graphicx}
\usepackage{booktabs}
\usepackage{colortbl}
\definecolor{grey1}{gray}{0.9}
\usepackage{multirow}
\usepackage{caption}
\usepackage{pifont}
\newcommand{\rot}[1]{\rotatebox{90}{#1}}
\newlength{\rotht}
\usepackage{breakcites}
\newcommand\samethanks[1][\value{footnote}]{\footnotemark[#1]}
\usepackage[accsupp]{axessibility}  

\usepackage{hyperref}

\usepackage{orcidlink}

\begin{document}

\title{Bootstrapping a 4D LiDAR Annotation Tool from Video Foundation Models} 

\titlerunning{LiDAR-SAM2}

\author{Jihun Kim\inst{1}\orcidlink{0009-0007-8764-195X}\thanks{Equal contribution.} \and
Hyun-Kurl Jang\inst{1}\orcidlink{0009-0003-7943-3326}\samethanks \and
Hyemin Yang\inst{1}\orcidlink{0009-0002-2112-7804}\samethanks \and
Jinnyeong Yang\inst{1}\orcidlink{0009-0002-9275-6296}\samethanks \and
Hyeokjun Kweon\inst{2}\orcidlink{0000-0003-4442-5513}\samethanks \and
 Kuk-Jin Yoon\inst{1}\orcidlink{0000-0002-1634-2756}}

\authorrunning{Kim et al.}

\institute{KAIST, Visual Intelligence Lab \\
\email{\{jihun1998, jhg0001, hyemin0806, jinnyeong6118,  kjyoon\}@kaist.ac.kr}\\
\and
Chung-Ang University, FoVLab\\
\email{hyeokjunkweon@cau.ac.kr}
}

\maketitle

\begin{center}
    \centering
    \captionsetup{type=figure}
    \vspace{-10pt}
    \includegraphics[width=.9\textwidth]{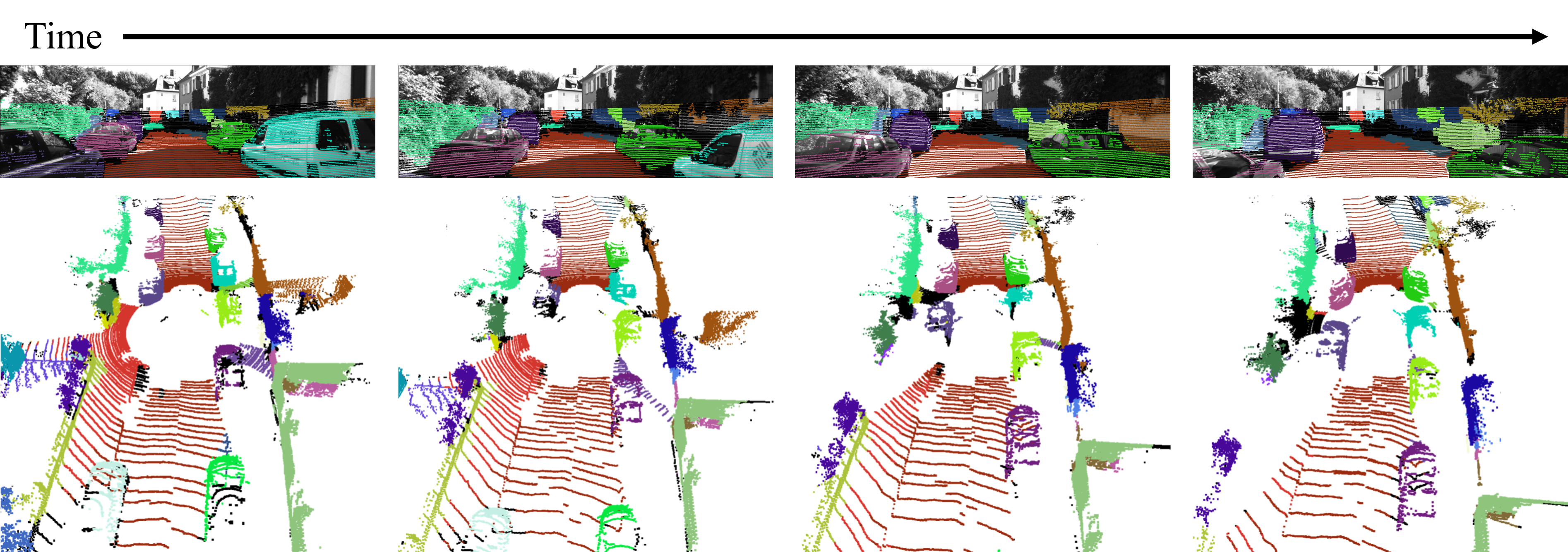}
    \vspace{-10pt}
    \caption{Segmentation results of LiDAR-SAM2 on SemanticKITTI~\cite{Dataset_ICCV2019_SEMANTICKITTI}. Each object is segmented via a user click in the first LiDAR frame, and distinct colors are assigned to different objects for visual distinction.}
    \vspace{-10pt}
    \label{fig:teaser}
\end{center}

\begin{abstract}
Progress in 4D LiDAR segmentation is bottlenecked by data.
Assigning temporally consistent labels across sparse point cloud sequences is
costly and hard to scale, and every new task or domain tends to demand fresh
dense annotation.
This motivates a simple question of whether high-quality LiDAR training data can
be produced automatically, without any human labeling.
To this end, we introduce \textbf{LiDAR-SAM2}, a framework that turns a 2D video
foundation model, SAM2, into a scalable source of supervision for the 4D LiDAR
domain.
On the data side, it automatically generates temporally coherent LiDAR-level
labels from SAM2 video masks through multi-view projection and spatio-temporal
aggregation.
On the modeling side, a tailored modality interface and a two-stage learning
objective adapt SAM2's video segmentation kernel to spatio-temporal LiDAR
structure, so that a single click per object yields a consistent mask track across
the sequence.
Trained with no human LiDAR annotation, LiDAR-SAM2 produces semantic and panoptic
labels on SemanticKITTI that approach the quality of full human annotation from
only a few points, and models trained on these labels approach the performance of
full ground-truth supervision.
This positions LiDAR-SAM2 as a scalable labeling tool that substantially reduces
the annotation burden for 3D and 4D scene understanding.
\end{abstract}

\section{Introduction}
\label{sec:intro}
Perception in autonomous driving demands a dynamic understanding of 3D scenes, and
\textbf{4D LiDAR segmentation} plays a pivotal role in this process, offering
spatially detailed interpretations of the scene by identifying objects and
temporally associating them across point cloud sequences.
The field covers a range of tasks, including
semantic~\cite{wang2025segnet4d,li2023memoryseg,liu2023mars3d,shi2024learning,wu2024taseg,shi2020spsequencenet,duerr2020lidar}, instance/panoptic~\cite{kreuzberg20224d,4dseg_ICRA2024_mask4former,marcuzzi2023mask4d,zhu20234d_eq4dpls,marcuzzi2022_CA_net,wang2023insmos,4dpanoptic}, and moving objects~\cite{chen2021moving,rozsa2025efficient,zeng2024mambamos} segmentation.
Each of these tasks relies on its own form of supervision, so that a separate set
of annotations must be collected for every task and granularity.

Obtaining such annotations directly on LiDAR point clouds, however, is
particularly difficult. Labeling sparse, irregular points is far less intuitive
than labeling images, and assigning temporally consistent labels across an entire
sequence makes the process even more labor-intensive and hard to scale. The
practical bottleneck of 4D LiDAR segmentation therefore lies less in the model than
in the data it depends on, suggesting the need to revisit the problem from a more
data-centric perspective.

Interestingly, a comparable challenge has arisen in the 2D domain.
Image segmentation, spanning semantic, panoptic, and instance formulations, has
suffered from the same rigidity of task definition and dependence on human
labeling.
To address these issues, a long line of research has explored interactive
segmentation, enabling users to guide models through visual prompts such as points.
This idea has recently culminated in foundation models such as
SAM~\cite{kirillov2023segment} and its video counterpart SAM2~\cite{ravi2024sam}.
By decoupling what to segment from how to segment, these models have achieved
remarkable scalability and generalization across object categories and visual
domains.
This paradigm has not only turned data annotation into an interactive process, but
has also established a foundation for general-purpose segmentation.

In light of these parallels, it is natural to ask whether the LiDAR domain could
benefit from a similar shift.
Recently, Interactive4D~\cite{fradlin2024interactive4d} takes a step in this
direction with a prompt-driven 4D LiDAR segmentation.
However, this approach still relies on human-annotated panoptic labels from a
specific dataset (SemanticKITTI~\cite{Dataset_ICCV2019_SEMANTICKITTI}).
As a result, it inherits the data limitations of prior task-specific approaches.
The segmentation granularity is fixed by the training annotations, and every new
domain would require re-collecting large amounts of manually labeled 4D data.

We propose \textbf{LiDAR-SAM2}, the first interactive 4D LiDAR segmentation
framework trained without any manually labeled LiDAR data.
Our central idea is to treat a 2D video foundation model as a scalable source of
LiDAR supervision rather than relying on a fixed, human-annotated dataset.
On the data side, LiDAR-SAM2 is trained entirely from pseudo-labels that are
automatically generated by SAM2 through multi-view projection and our
spatio-temporal 4D mask aggregation, which turns partial per-camera masks into
dense, temporally coherent point-wise labels.
On the structural and learning side, we design a tailored modality conversion
interface and a two-stage learning objective that adapt SAM2's RGB video
segmentation kernel to 4D LiDAR geometry and temporal structure.

Across extensive experiments, LiDAR-SAM2 shows (1) strong labeling quality on
SemanticKITTI~\cite{Dataset_ICCV2019_SEMANTICKITTI}, producing semantic and
panoptic supervision without any human-annotated LiDAR labels, and (2) high
downstream utility, where models trained on our automatically generated labels
approach the accuracy of models trained on full manual annotation from only
minimal prompts.
Figure~\ref{fig:teaser} further illustrates the resulting interactive 4D LiDAR
segmentation, where a single initial click per object enables LiDAR-SAM2 to produce
consistent, high-quality mask tracks across the entire sequence.

Together, these results show that better training data, curated automatically from
a foundation model, can bring the scalability and generality of interactive
segmentation into the 4D LiDAR domain, offering a practical path toward
large-scale, high-quality 3D scene understanding without manual annotation.

\section{Related Works}
\label{sec:rw}

\subsection{4D LiDAR Segmentation}
LiDAR segmentation is fundamental to understanding dynamic 3D environment, considering temporal coherence across sequences.
Early approaches employed point-based or voxelized backbones~\cite{thomas2019kpconv, zhu2021cylindrical, zhou2020cylinder3d, choy20194d} to model spatial geometry within a single scan, while later 4D methods incorporated temporal cues through recurrent units, temporal attention, or pose-aware feature warping~\cite{4dpanoptic, marcuzzi2023mask4d, kreuzberg20224d, mersch2022receding, wang2025segnet4d, 4dseg_ICRA2024_mask4former}.
The field now covers diverse tasks—including semantic, instance, panoptic, and moving objects segmentation—each imposing different levels of label granularity and temporal consistency~\cite{4dpanoptic, wang2025segnet4d, mersch2022receding}.
Despite these advances, most systems still depend on manually annotated point-level labels, which are expensive and labor-intensive to collect at scale.
To address this challenge, we propose an interactive 4D LiDAR segmentation named LiDAR-SAM2, as a labeling assistant.
It produces high-quality spatio-temporal segmentation from minimal point prompts, significantly reducing the annotation burden, and is broadly applicable across downstream LiDAR segmentation tasks.

\subsection{Leveraging 2D VFMs for 3D}
Vision foundation models (VFMs) have shown strong transfer to downstream tasks, with especially large gains in 2D settings where pretraining is image/video-centric. Extending 2D VFMs to 3D perception has proceeded along two primary routes. Lifting-based methods \cite{2D3Dproj_partslip, 2D3Dproj_partslip++, 2D3Dproj_ICCVw2023_sam3d, 2D3Dproj_neurips2025_sa3dip, 2D3Dproj_SA3D, 2D3Dproj_sai3d, 2D3Dproj_pointseg,kweon2024weakly} first obtain 2D predictions from a VFM and then reproject them into 3D using calibrated geometry and visibility reasoning. Distillation-based \cite{2D3Dproj_IROS2025_labelefficientLPS, zhou2024pointsampromptable3dsegmentation,2024_SAL, 2025sal4d, 2D3Dproj_ICCV2025_sam4d, 2D3Dproj_partfield, 2D3Dproj_sampart3d} approaches instead train 3D-native backbones under supervision from 2D VFMs via pseudo-labels or feature-level guidance.
Both strategies, however, have limitations: lifting pipelines are brittle to calibration and resampling and often underutilize native 3D geometry, while distillation often erode advantages of large-scale 2D VFM—broad category coverage and open-set generalization. Complementary to these, another line of work incorporates 2D VFMs within 3D frameworks by adding 3D-aware adapters for volumetric inputs~\cite{wang2022p2p,2D3Dproj_autoprosam, li2025ga}. We advance this line in the LiDAR domain by adapting SAM2 via a geometry-preserving interface that enables promptable, LiDAR-only inference.

\section{Pseudo-Label Generation using SAM2}\label{sec:plg}

We begin by describing how to generate 4D pseudo-labels from synchronized multi-view RGB images. 
Specifically, SAM2~\cite{ravi2024sam} is applied to the multi-view RGB streams to obtain per-view temporal segmentation masks, which are then transferred to the LiDAR domain via geometric projection and multi-view aggregation. 
This produces temporally coherent point-wise pseudo-labels across the LiDAR sequence, enabling us to train an interactive 4D LiDAR segmentation model without any manually annotated 3D data. 

\subsection{Obtaining View-wise Mask Proposals}\label{sec:obtaining_view_wise_mask}

We denote a time-ordered LiDAR sequence by
\begin{equation}
\mathcal{X} = \{ X_{t} \}_{t=1}^{T}, \quad
X_{t} = \{ x_{t,i} \in \mathbb{R}^3 \}_{i=1}^{N_t},
\end{equation}
where $T$ is the length of the sequence and $N_t$ is the number of points in frame $t$.
We assume access to a synchronized RGB image $I_{t} \in \mathbb{R}^{H_I \times W_I \times 3}$ at each timestep, with known geometric calibration between the LiDAR and the camera.

We exploit the segment-everything mode\footnote{The segment everything mode is implemented in SAM but not yet in SAM2; we thus use SAM for this step without methodological differences.} to obtain initial mask proposals.
This yields a set of 2D binary masks $\{\, m_{1,k}^{2D} \,\}_{k=1}^{K_1}$, where $m_{1,k}^{2D} \in \{0,1\}^{H_I \times W_I}$.
$K_1$ denotes the number of proposals in the initial frame.

Subsequently, each mask $m_{1,k}^{2D}$ is then used as a prompt to SAM2~\cite{ravi2024sam} to obtain its temporal mask trajectory:
\begin{equation}
\mathrm{SAM2}(V \mid m_{1,k}^{2D} ) \;\;\;\rightarrow\;\;\; \mathcal{M}_k^{Video}=\{m_{t,k}^{2D}\}^T_{t=1}
\end{equation}
where $V=I_{1:T}$ denotes the video (\textit{i.e.}, image sequence).
In other words, SAM2 propagates each initial mask proposal forward through time, producing a consistent segmentation trajectory for each segment across the sequence.

Then, each 2D binary mask is transferred into the 3D LiDAR domain via the calibrated LiDAR-camera projection.
We use a lifting function $\Phi_t : \{0,1\}^{H_I \times W_I} \rightarrow \{0,1\}^{N_t}$ to obtain 3D binary mask of $k$-th proposal at timestep $t$ as
\begin{equation}
m_{t,k}^{3D} = \Phi_t(m_{t,k}^{2D}),\quad
m_{t,k}^{3D} \in \{0,1\}^{N_t}.
\end{equation}
Collectively performing this process on the video mask $\mathcal{M}_k^{Video}$ yields a 4D LiDAR mask track:
\begin{equation}
\mathcal{M}^{4D}_k = \{\, m_{t,k}^{3D} \,\}_{t=1}^{T}.
\end{equation}

\subsection{Spatio-Temporal 4D Mask Aggregation}

The above process assumes a single view per LiDAR scan, but practical setups often use multiple cameras. 
While each camera yields temporally consistent mask tracks, cross-view consistency is not guaranteed. 
With limited camera coverage (\textit{e.g.} SemanticKITTI~\cite{Dataset_ICCV2019_SEMANTICKITTI}), each view captures only a subset of the scene, causing the same object to split into multiple view-specific mask tracks.

To obtain a unified set of 4D pseudo-labels in the LiDAR domain, we perform spatio-temporal 4D mask aggregation.
It aims to merge view-specific tracks that correspond to the same underlying 3D object and to stitch identities over time despite partial view coverage.
We begin with spatial aggregation. 
For each object, multiple view-specific masks may correspond to the same physical instance. 
To identify such correspondences, we assign the same mask track ID to any set of masks that share common LiDAR points and merge them accordingly.
Applying this rule across all views results in a cross-view consistent set of mask tracks.

We then perform temporal aggregation to enforce consistency across time. Since 2D masks are obtained from camera views, the initial pseudo-labels are inherently limited to camera-visible regions, leaving portions of the LiDAR sweep unlabeled due to the restricted camera FoV. To overcome this limitation, we explicitly expand pseudo-label coverage via spatiotemporal aggregation. Concretely, masks from the previous timestep are first propagated into the current LiDAR frame using the known ego-motion. We then robustly fuse the propagated and current masks by voxelizing the point cloud and assigning each voxel the majority mask ID among the points it contains; the voted IDs are broadcasted back to points. Iterating this propagation-and-fusion over time allows labels to persist and grow beyond what is labeled in the current frame alone, mitigating the limited camera coverage. More details are in the Supplementary Material.

On SemanticKITTI, temporal aggregation substantially improves pseudo-label density while preserving quality. Without aggregation, we recover 22.12 segments on average, achieving 91.90\% precision but only 13.59\% recall. After aggregation, the average number of segments increases to 73.84 and recall rises dramatically to 79.58\%, while precision remains high at 85.93\%. This yields temporally consistent dense pseudo-labels, enabling high-quality automatic supervision.

\begin{figure*}[t]
    \centering
    \includegraphics[width=0.99\linewidth]{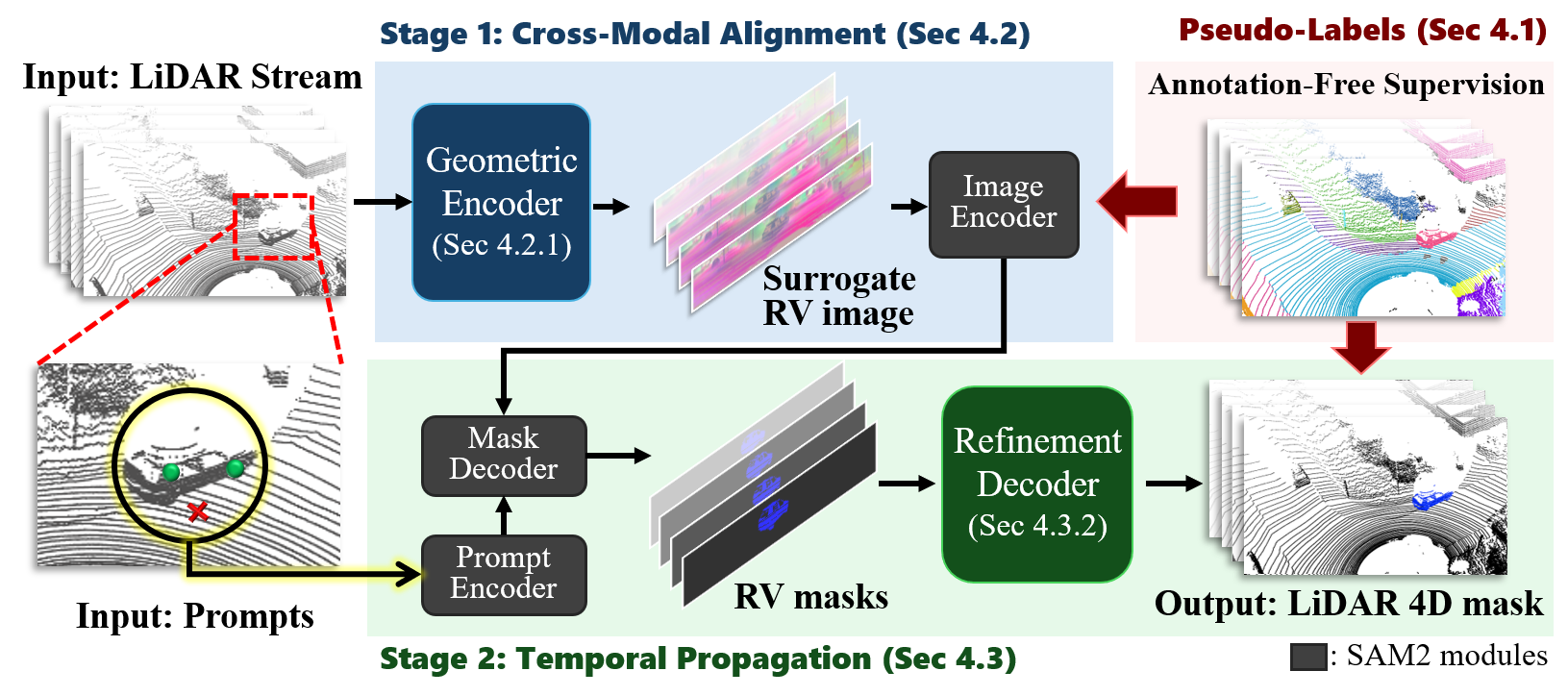}
    \vspace{-10pt}
    \caption{Overview of the LiDAR-SAM2 framework. The training pipeline consists of two stages, both supervised by pseudo-labels.}
    \vspace{-10pt}
    \label{fig:Main}
\end{figure*}

\section{LiDAR-SAM2}\label{sec:method}

\subsection{Overview}

Our goal is to build an interactive 4D LiDAR segmentation model that allows a user to specify what to segment through a minimal prompt. 
Following SAM2~\cite{ravi2024sam}, we aim to learn 
\begin{equation}
\hat{\mathcal{Y}} = \mathcal{F}(\mathcal{X} \mid \mathcal{P}), \quad \hat{\mathcal{Y}} = \{ \hat{Y}_t \}_{t=1}^{T},
\label{eq:method-def}
\end{equation}
where $\mathcal{X} = \{ X_t \}_{t=1}^{T}$ is the LiDAR sequence and $\mathcal{P} \subset X_{t}$ is a set of user-selected point prompts.
The output $\hat{\mathcal{Y}}$ is a predicted 4D mask track, where $\hat{Y}_{t} = \{ \hat{y}_{t,i} \in [0,1] \}_{i=1}^{N_t}$
denotes per-point segmentation scores indicating how likely each point $x_{t,i}$ belongs to the segment specified by $\mathcal{P}$.

On the architectural side, rather than designing a new model from scratch, we retain SAM2 as a strong segmentation kernel, and focus on enabling it to operate directly on LiDAR sequences.
This preserves SAM2’s interactive prompt-based behavior, but also introduces two challenges:
(1) LiDAR point clouds are sparse and irregular 3D data, whereas SAM2 expects dense 2D feature maps;
and (2) the temporal dynamics of 4D LiDAR sequences differ from those in videos on which SAM2 was originally trained.

To address these issues, we propose an interface-oriented architecture that maps LiDAR frames into and out of SAM2’s 2D representation space while preserving their underlying 3D geometric structure.
This allows SAM2 to process LiDAR inputs without modifying its architecture, maintaining spatial locality and structure in point space.

We further introduce a two-stage strategy with specialized training objectives.
This design is inspired by recent VLMs~\cite{llava, blip2}, where the vision-to-language projection layer is first aligned before full fine-tuning.
In \textbf{Stage~1}, we perform cross-modal alignment so that SAM2’s image encoder can effectively interpret LiDAR-derived range-view representations. This alignment is conducted at the frame level, without any temporal modeling.
Then, in \textbf{Stage~2}, we enable LiDAR-SAM2 to learn temporal propagation and object-level consistency across time for 4D segmentation.

\subsection{Learning Cross-Modal Alignment (Stage 1)}

The goal of Stage~1 is to align LiDAR domain with SAM2 in a \textbf{frame-wise} manner. 
Directly learning temporal propagation and motion consistency in 4D space is challenging, and SAM2 is trained on the RGB domain only.
Therefore, we first ensure that a single LiDAR scan and its information can be interpreted by SAM2.
For clarity, since Stage~1 is frame-wise, we omit the time index $t$ in this section.

We adopt the Range-View (RV) representation to express each LiDAR scan as a dense image-like signal, allowing SAM2 to be used without modifying its architecture.
$\rho(\cdot)$ denotes the standard spherical projection from 3D space to RV plane, mapping the points $x_i \in \mathbb{R}^3$ to pixel coordinates $(u,v) \in \{1,\dots,H\}\times\{1,\dots,W\}$.

In addition, the 4D pseudo-labels from Sec.~\ref{sec:plg} are also projected into the RV domain.
Given a mask track $\mathcal{M}^{4D}_k = \{m^{3D}_{t,k}\}_{t=1}^{T}$, each per-frame binary point mask $m^{3D}_{k} \in \{0,1\}^{N}$ is mapped to a range-view binary image:
\begin{equation}
m^{RV}_{k} = \rho(m^{3D}_{k}) \in \{0,1\}^{H\times W},
\label{eq:mask_rv}
\end{equation}
which is spatially aligned with the RV domain.

\subsubsection{Geometry-aware Range-View Representation}

To preserve geometric structure while converting LiDAR into an image-like representation, we first extract per-point features using a SSL-pretrained geometric encoder $E_{\mathrm{geo}}$:
\begin{equation}
\mathbf{h} = E_{\mathrm{geo}}(X) \in \mathbb{R}^{N \times d}.
\end{equation}

To make these features compatible with SAM2’s image encoder trained on the RGB domain, we employ a lightweight MLP $g$ to map them into a 3-channel embedding:
\begin{equation}
\mathbf{c} = g(\mathbf{h}) \in \mathbb{R}^{N \times 3},
\end{equation}
    interpreting LiDAR features as an RGB signal.
We then rasterize the point-wise embeddings into a dense RV image:
\begin{equation}
\hat{I}^\text{RV} = \rho(X, \mathbf{c}) \in \mathbb{R}^{H\times W \times 3}.
\end{equation}

The resulting surrogate RV image is processed by the SAM2 image encoder with trainable LoRA~\cite{hu2022lora}:
\begin{equation}
\mathbf{F} = E_{\mathrm{SAM2}}(\hat{I}^\text{RV}) \in \mathbb{R}^{H \times W \times D}.
\end{equation}

\subsubsection{Objective Function for Interactive Segmentation}

We train our framework with SAM2-style loss.
From the pseudo-label $m^{RV}$, we sample a set of foreground/background pixels to serve as positive/negative prompts.
This set of prompts $\mathcal{P}$ is fed into the prompt encoder, and combined with the feature map $\mathbf{F}$ through the SAM2 decoder to predict a mask $\hat{y}^{RV}$.
Accordingly, the loss is defined by a segmentation loss between the predicted mask and the pseudo-label in the RV domain: \begin{equation}
\mathcal{L}^{\text{frame-wise}}_{\text{interactive}}
=
\text{RVSegLoss}(\hat{y}^{RV}, m^{RV}),
\label{eq:rvsegloss}
\end{equation}
where the details about RVSegLoss is in \textit{Supplementary Material}.
To sum up, the full objective for Stage~1 is:
\begin{equation}
\mathcal{L}_{\text{Stage1}}
=
\mathcal{L}^{\text{frame-wise}}_{\text{interactive}}.
\label{eq:loss_stage1_all}
\end{equation}
Note that Stage~1 updates the LiDAR encoder ($E_\text{geo}$), the lightweight MLP ($g$), and SAM2's image encoder ($E_\text{SAM2}$ with LoRA).
The other parts of SAM2 remain unchanged.

\subsection{Learning Temporal Propagation (Stage 2)}

Stage~2 focuses on learning how segmentation should propagate over time.  
The model receives a LiDAR sequence $\mathcal{X}=\{X_t\}_{t=1}^{T}$, and each scan is converted into its RV feature map: $\mathbf{F}_t = E_{\mathrm{SAM2}}(\mathrm{RV}_t)$.

\subsubsection{4D Decoder for Refinement} \label{sec:superimpose}
Since this track is still in the RV domain, we obtain the final 4D point-wise mask track by inverse projection:
\begin{equation}
\hat{\mathcal{Y}}^{4D} = \{\hat{y}^{4D}_{t}\}_{t=1}^T, \quad \hat{y}^{4D}_{t}= \rho^{-1}(\hat{y}^{RV}_t).
\label{eq:stage2-lift}
\end{equation}
To further restore LiDAR-level geometric detail, we introduce a lightweight 3D refinement LiDAR decoder $D_{\mathrm{ref}}$.
Specifically, given a temporal window $t : t+\Delta t$, we place the lifted predictions\footnote{In practice, we use the lifted SAM2 decoder logits before thresholding rather than binary masks. Details can be found in Supplementary Material.} into $D_{\mathrm{ref}}$ for each frame in the window and refine them in point space. We denote this refinement over the entire window as
\begin{equation}
\tilde{\mathcal{Y}}^{4D}_{t:t+\Delta t}
=
D_{\mathrm{ref}}(X_{t:t+\Delta t}, \hat{\mathcal{Y}}^{4D}_{t:t+\Delta t}).
\label{eq:refine}
\end{equation}
This decoding stage corrects projection-induced artifacts and reinstates surface continuity in 3D space, yielding coherent and stable segmentation results.

Finally, the refined 4D mask track $\tilde{\mathcal{Y}}^{4D}$ is supervised using the 4D pseudo-label $\mathcal{M}^{4D}$ in the LiDAR domain via
\begin{equation}
\mathcal{L}_{\text{refine}}
=
\text{4DSegLoss}(\tilde{\mathcal{Y}}^{4D},\mathcal{M}^{4D}),
\label{eq:4dsegloss}
\end{equation}
where the details about 4DSegLoss are in \textit{Supplementary Material}.
To sum up, the full objective for Stage~2 is:
\begin{equation}
\mathcal{L}_{\text{Stage2}}
=
\mathcal{L}^{\text{temporal}}_{\text{interactive}}
+
\mathcal{L}_{\text{refine}}.
\label{eq:loss_stage2_all}
\end{equation}

During Stage~2, we freeze the modules trained in Stage~1 ($E_{\mathrm{geo}}$, $g$, $E_{\mathrm{SAM2}}$), and train the remaining SAM2 components (prompt encoder, memory attention, and mask decoder) as well as the additional temporal modules introduced below.
This allows the model to learn how to propagate segmentation consistently across time without disrupting the modality alignment established in Stage~1.

\section{Experiments}

\subsection{Settings}
\noindent\textbf{Datasets}
We evaluate LiDAR-SAM2 on SemanticKITTI~\cite{Dataset_ICCV2019_SEMANTICKITTI},
which consists of 64-beam LiDAR scans synchronized with two forward-facing RGB
cameras. We train on the training set and report all results on the
validation set.

\noindent\textbf{Metric}
For semantic segmentation, we report the mean Intersection-over-Union (mIoU) averaged over the 19 SemanticKITTI classes. For 4D panoptic segmentation, we adopt the LiDAR Segmentation and Tracking Quality (LSTQ)~\cite{4dpanoptic}, which factorizes into an association score ($S_{assoc}$) measuring temporal instance consistency and a classification score ($S_{cls}$). We further report semantic IoU separately for stuff ($IoU_{st}$) and thing ($IoU_{th}$) classes. All metrics are computed on the respective validation sets.

\noindent\textbf{Implementation}
All experiments are conducted on a single NVIDIA A6000 GPU with the pretrained SAM2 \emph{base+}. 
The geometric encoder is pre-trained using FCGF~\cite{choy2019fully} on the same datasets used for evaluation.
The refinement head $D_{\mathrm{ref}}$ is implemented as MinkUNet-14~\cite{choy20194d}. 

\subsection{LiDAR-SAM2 as Labeling Tool}

We evaluate LiDAR-SAM2 as a labeling tool for all entities in the LiDAR scene by
simulating human interaction on 4-frame LiDAR sweeps.
At the first LiDAR frame, a fixed number of point prompts is provided for every
instance to initialize the annotation, following a standard interactive
segmentation protocol.
The first positive click is placed near the object center, and subsequent clicks
are added by inspecting the true-negative and false-positive regions, where we
click on whichever region is larger at that time.
For stuff categories such as road and sidewalk that are not instance-separated, we
voxelize the current point cloud with a voxel size of 0.2\,m and compute connected
components in voxel space, treating each disconnected component as a separate
pseudo-instance even when several share the same semantic label, which keeps
propagation and editing localized and consistent across sweeps.
The predicted masks are then propagated through the remaining frames and one
additional frame to seed the next sweep.
Carried masks are reused in the following sweep only if their overlap with the
corresponding ground-truth mask exceeds an IoU threshold of 0.8, and otherwise,
including for newly appearing objects, we discard the carried mask and introduce
additional human prompts to re-initialize the instance.
On average, this process corresponds to roughly \textbf{ten initial point prompts per frame}.

\subsubsection{Semantic Segmentation}
\label{sec:semseg}

We first assess our labels on semantic segmentation by training two standard backbones, MinkowskiUNet~(MinkUNet)~\cite{choy20194d} and PointTransformer v2~(PTv2)~\cite{wu2022point}, from scratch on labels of different origin. As reported in Table~\ref{tab:annotation_semseg}, we compare four label sources. The first is the raw point prompts used during interaction. 
The second is labels from SAM2 inference via RGB projection. The third is labels generated by LiDAR-SAM2. The fourth is full human ground truth. 
Training on the sparse point prompts alone yields very low accuracy, reflecting the limited supervision. SAM2 labels give a moderate improvement but remain insufficient, suffering from projection artifacts and a lack of adaptation to the LiDAR domain. 
In contrast, LiDAR-SAM2 labels lead to a substantial gain, roughly doubling the mIoU of naive SAM2, and bring both backbones close to models trained on full ground truth, despite using no human LiDAR annotation.

\begin{table}[t]
    \centering
    \caption{Semantic segmentation results (mIoU) trained with point labels, SAM2 labels, our labels, and full GT. Experiments are on the SemanticKITTI~\cite{Dataset_ICCV2019_SEMANTICKITTI} validation set.}
    \resizebox{\linewidth}{!}{
    \begin{tabular}{l|l|ccccccccccccccccccc|c}
        \toprule
       \raisebox{0.4\rotht}{Methods} & \raisebox{0.4\rotht}{Label type} & \rot{car}                  &                 \rot{bicycle}                   &         \rot{motorcycle}                            &                    \rot{truck}                   &                \rot{other-vehicle}                    &             \rot{person}                    &           \rot{bicyclist}                            &      \rot{motorcyclist}                            &                     \rot{road}              &              \rot{parking}                    &                \rot{sidewalk}                       &                      \rot{other-ground}             &       \rot{building}   &       \rot{fence}  &       \rot{vegetation}   &       \rot{trunk}  &       \rot{terrain}  &       \rot{pole}  &       \rot{traffic-sign}    &               \rot{mIoU}                      \\  
        \hline\hline
       \multirow{4}{*}{MinkU~\cite{choy20194d}} &  Point labels & 9.7 & 0.5 & 2.4 & 5.6 & 4.0 & 0.8 & 1.6 & 3.4 & 0.8 & 2.5 & 4.1 & 0.2 & 5.3 & 4.0 & 1.8 & 10.4 & 10.4 & 23.1 & 7.4 & 5.2  \\ 
        & SAM2 labels   &  61.0 & 0.6 & 6.5 & 50.3 & 11.9 & 25.1 & 44.7 & 0.1 & 57.8 & 6.0 & 7.4 & 0.6 & 67.6 & 21.8 & 53.5 & 21.3 & 47.8 & 45.7 & 37.2 &  29.8 \\
       & Our labels   &  84.3 & 17.8 & 60.5 & 75.8 & 46.8 & 58.3 & 73.0 & 2.5 & 85.2 & 36.8 & 69.6 & 0.3 & 84.6 & 52.2 & 84.7 & 63.6 & 69.9 & 49.6 & 33.2 & 55.2   \\
       &     Full GT   & 96.1 & 35.0 & 64.8 & 83.4 & 59.9 & 73.4 & 88.3 & 0.0 & 93.7 & 53.1 & 81.0 & 7.0 & 91.0 & 61.7 & 88.0 & 67.8 & 75.5 & 62.5 & 48.6 & 64.8  \\
        \hline
       \multirow{4}{*}{PTv2~\cite{wu2022point}} &  Point labels  &  27.6 & 1.9 & 5.8 & 3.2 & 6.6 & 0.9 & 2.0 & 0.0 & 1.2 & 4.8 & 0.2 & 0.1 & 6.2 & 3.7 & 1.8 & 15.0 & 2.7 & 41.0 & 16.6  & 7.4   \\ 
        & SAM2 labels & 78.7 & 0.1 & 0.0 & 0.0 & 15.3 & 14.6 & 22.3 & 1.9 & 53.9 & 1.4 & 0.7 & 0.0 & 72.1 & 23.1 & 64.5 & 22.8 & 51.5 & 43.2 & 35.9  & 26.4  \\
       & Our labels  & 86.2 & 34.1 & 71.9 & 78.2 & 39.7 & 69.1 & 80.7 & 30.1 & 86.6 & 45.1 & 72.9 & 8.3 & 86.3 & 59.4 & 86.8 & 62.2 & 73.7 & 52.9 & 41.1 & 61.3  \\
       &     Full GT  & 96.0 & 53.5 & 83.9 & 88.6 & 59.4 & 82.2 & 93.4 & 19.2 & 95.3 & 48.7 & 83.2 & 0.1 &91.3  & 65.1 & 88.1 & 71.2 & 74.3  & 65.9 & 50.8  & 69.0  \\
        \bottomrule
        \end{tabular}
        \label{tab:annotation_semseg}
        }
        \vspace{5pt}
\end{table}

\subsubsection{Panoptic Segmentation}
\label{sec:panseg}
We further validate our labels on 4D panoptic segmentation, which additionally
requires temporally consistent instance identities. We train three panoptic
backbones, namely 4D-PLS~\cite{4dpanoptic}, 4D-StOP~\cite{kreuzberg20224d}, and
Mask4Former~\cite{4dseg_ICRA2024_mask4former}, using labels generated by LiDAR-SAM2, and compare
against the same backbones trained on full ground truth (Table~\ref{tab:annotation_panseg}).
Models supervised by our labels achieve strong panoptic quality across all
metrics, including LSTQ, $S_\text{assoc}$, $S_\text{cls}$, and the stuff and thing
IoUs. They recover a large fraction of the full ground-truth performance without
any manual LiDAR labels.

\begin{table}[t]
    \centering
    \caption{Panoptic segmentation results trained with our labels and full GT. Experiments are evaluated on SemanticKITTI~\cite{Dataset_ICCV2019_SEMANTICKITTI} validation set.}
    \resizebox{0.7\linewidth}{!}{
    \begin{tabular}{l|l|ccccc}
    \toprule
     Methods & Label type & LSTQ & $S_{assoc}$ & $S_{cls}$ & IoU$_{st}$ & IoU$_{th}$ \\
    \hline\hline
    4D-PLS~\cite{4dpanoptic} & Our labels & 46.9 & 46.3  & 47.6 & 54.3 & 44.3 \\
    4D-PLS~\cite{4dpanoptic} & Full GT & 62.7 & 65.1 & 60.5 & 65.4 & 61.3 \\\hline
    4D-StOP~\cite{kreuzberg20224d} & Our labels & 57.7 & 64.2  & 51.7 & 56.5 & 51.7 \\
    4D-StOP~\cite{kreuzberg20224d} & Full GT & 67.0 & 74.4 & 60.3 & 65.3 & 60.9 \\\hline
    Mask4Former~\cite{4dseg_ICRA2024_mask4former} & Our labels & 64.8 & 70.7  & 59.4 & 56.9 & 62.8 \\
    Mask4Former~\cite{4dseg_ICRA2024_mask4former} & Full GT & 70.5 & 74.3 & 66.9 & 67.1 & 66.6 \\
    \bottomrule
    \end{tabular}
    \label{tab:annotation_panseg}
    }
\end{table}

\subsubsection{Labeling Quality}
\label{sec:labelqual}
\begin{figure}[t]
    \centering
    \includegraphics[width=0.99 \linewidth]{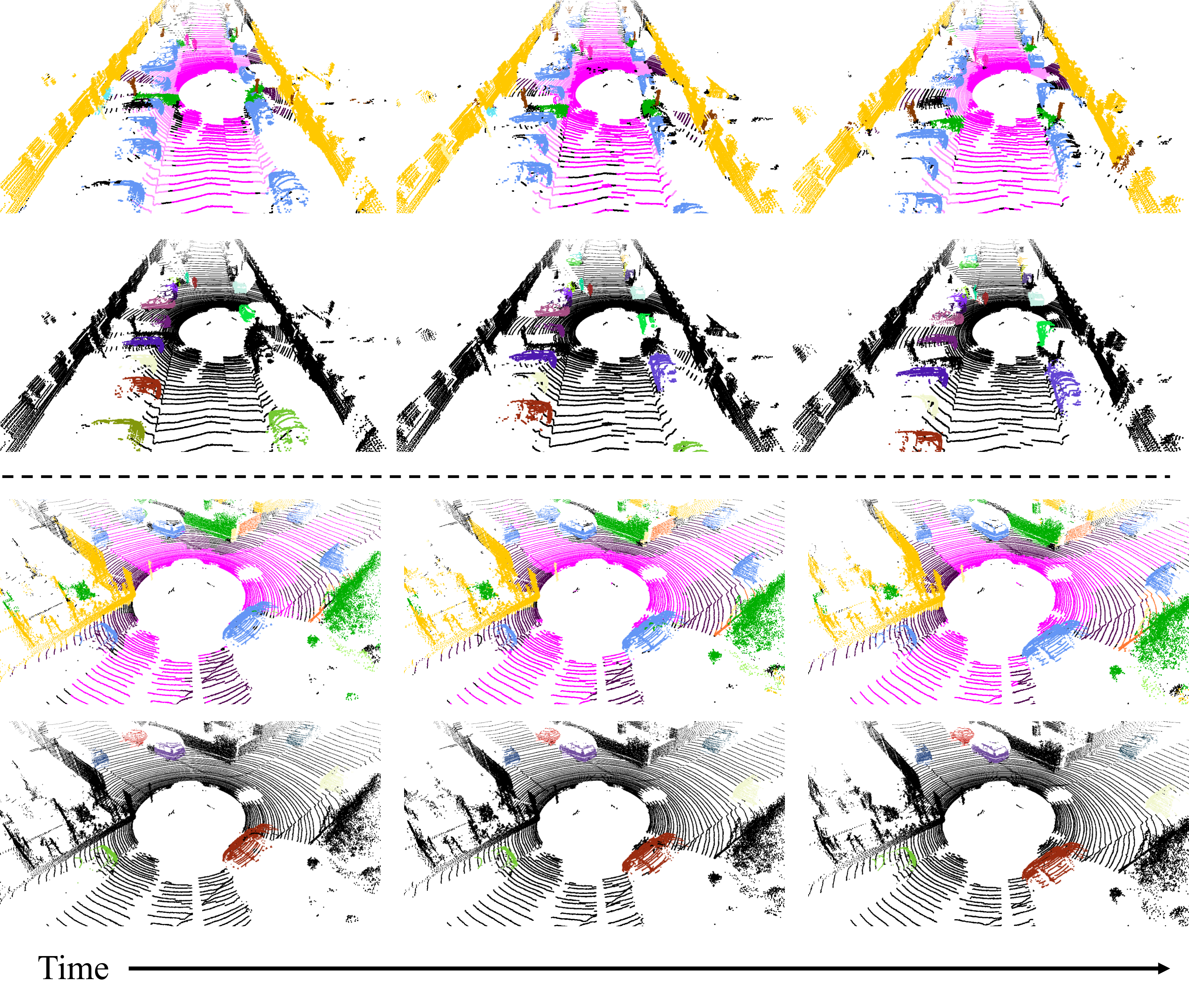}
    \vspace{-13pt}
    \caption{Qualitative results of labels annotated with LiDAR-SAM2. The top shows instance labels, while the bottom shows semantic labels.}
    \label{fig:annotation}
    \vspace{-15pt}
\end{figure}

Beyond downstream accuracy, we inspect the labels produced by LiDAR-SAM2 directly.
Figure~\ref{fig:annotation} shows instance-level and semantic-level labels generated by our
tool. Object boundaries are clean and identities remain consistent across time,
closely matching the underlying scene structure. Together with the strong
downstream performance reported above, these labels confirm that LiDAR-SAM2 is a
readily deployable labeling tool for 3D and 4D LiDAR segmentation, dramatically
reducing the manual annotation burden.

\vspace{-5pt}
\section{Conclusion}
We present LiDAR-SAM2, a novel framework for interactive 4D LiDAR segmentation that requires no manual LiDAR labels.
By distilling temporal priors from SAM2 into the LiDAR domain, our approach leverages synchronized video–LiDAR sequences to train a LiDAR-only model capable of producing spatiotemporally consistent interactive segmentation.
A tailored LiDAR-aware pipeline and specialized training objectives enable SAM2’s video segmentation kernel to operate effectively on LiDAR data while preserving strong temporal coherence.
Our experiments demonstrate that LiDAR-SAM2 substantially reduces annotation cost, achieving strong performance with only minimal prompting.

\section{Acknowledgement}

This work was supported by the Ministry of Education of the Republic of Korea and the National Research Foundation of Korea (NRF-2025S1A5C3A04022639).



%
%
\bibliographystyle{splncs04}
\bibliography{main}

@String(CVPR  = {IEEE Conf. Comput. Vis. Pattern Recog.})

@String(ICCV  = {Int. Conf. Comput. Vis.})

@String(ICLR  = {Int. Conf. Learn. Represent.})

@String(CVPR  = {CVPR})

@String(ICCV  = {ICCV})

@String(ICLR  = {ICLR})

@article{fradlin2024interactive4d,
  title={Interactive4d: Interactive 4d lidar segmentation},
  author={Fradlin, Ilya and Zulfikar, Idil Esen and Yilmaz, Kadir and Kontogianni, Theodora and Leibe, Bastian},
  journal={arXiv preprint arXiv:2410.08206},
  year={2024}
}

@inproceedings{2024_SAL,
  title={Better call sal: Towards learning to segment anything in lidar},
  author={O{\v{s}}ep, Aljo{\v{s}}a and Meinhardt, Tim and Ferroni, Francesco and Peri, Neehar and Ramanan, Deva and Leal-Taix{\'e}, Laura},
  booktitle={European Conference on Computer Vision},
  pages={71--90},
  year={2024},
  organization={Springer}
}

@inproceedings{2025sal4d,
    title={{Zero-Shot 4D Lidar Panoptic Segmentation}},
    author={Zhang, Yushan and O\v{s}ep, Aljo\v{s}a and Leal-Taix\'{e}, Laura and Meinhardt, Tim},
    booktitle={Conference on Computer Vision and Pattern Recognition (CVPR)},
    year={2025},
}

@inproceedings{4dpanoptic,
  title={4d panoptic lidar segmentation},
  author={Aygun, Mehmet and Osep, Aljosa and Weber, Mark and Maximov, Maxim and Stachniss, Cyrill and Behley, Jens and Leal-Taix{\'e}, Laura},
  booktitle={Proceedings of the IEEE/CVF Conference on Computer Vision and Pattern Recognition},
  pages={5527--5537},
  year={2021}
}

@inproceedings{zhou2024pointsampromptable3dsegmentation,
      title={Point-SAM: Promptable 3D Segmentation Model for Point Clouds}, 
      author={Yuchen Zhou and Jiayuan Gu and Tung Yen Chiang and Fanbo Xiang and Hao Su},
      year={2024},
      eprint={2406.17741},
      archivePrefix={arXiv},
      primaryClass={cs.CV}
}

@article{wang2022p2p,
  title={P2p: Tuning pre-trained image models for point cloud analysis with point-to-pixel prompting},
  author={Wang, Ziyi and Yu, Xumin and Rao, Yongming and Zhou, Jie and Lu, Jiwen},
  journal={Advances in neural information processing systems},
  volume={35},
  pages={14388--14402},
  year={2022}
}

@article{ravi2024sam,
  title={Sam 2: Segment anything in images and videos},
  author={Ravi, Nikhila and Gabeur, Valentin and Hu, Yuan-Ting and Hu, Ronghang and Ryali, Chaitanya and Ma, Tengyu and Khedr, Haitham and R{\"a}dle, Roman and Rolland, Chloe and Gustafson, Laura and others},
  journal={arXiv preprint arXiv:2408.00714},
  year={2024}
}

@inproceedings{kirillov2023segment,
  title={Segment anything},
  author={Kirillov, Alexander and Mintun, Eric and Ravi, Nikhila and Mao, Hanzi and Rolland, Chloe and Gustafson, Laura and Xiao, Tete and Whitehead, Spencer and Berg, Alexander C and Lo, Wan-Yen and others},
  booktitle={Proceedings of the IEEE/CVF international conference on computer vision},
  pages={4015--4026},
  year={2023}
}

@inproceedings{Dataset_ICCV2019_SEMANTICKITTI,
  author = {J. Behley and M. Garbade and A. Milioto and J. Quenzel and S. Behnke and C. Stachniss and J. Gall},
  title = {{SemanticKITTI: A Dataset for Semantic Scene Understanding of LiDAR Sequences}},
  booktitle = {Proc. of the IEEE/CVF International Conf.~on Computer Vision (ICCV)},
  year = {2019}
}

@inproceedings{4dseg_ICRA2024_mask4former,
  title={Mask4former: Mask transformer for 4d panoptic segmentation},
  author={Yilmaz, Kadir and Schult, Jonas and Nekrasov, Alexey and Leibe, Bastian},
  booktitle={2024 IEEE International Conference on Robotics and Automation (ICRA)},
  pages={9418--9425},
  year={2024},
  organization={IEEE}
}

@article{wu2022point,
  title={Point transformer v2: Grouped vector attention and partition-based pooling},
  author={Wu, Xiaoyang and Lao, Yixing and Jiang, Li and Liu, Xihui and Zhao, Hengshuang},
  journal={Advances in Neural Information Processing Systems},
  volume={35},
  pages={33330--33342},
  year={2022}
}

@inproceedings{choy20194d,
  title={4d spatio-temporal convnets: Minkowski convolutional neural networks},
  author={Choy, Christopher and Gwak, JunYoung and Savarese, Silvio},
  booktitle={Proceedings of the IEEE/CVF conference on computer vision and pattern recognition},
  pages={3075--3084},
  year={2019}
}

@inproceedings{choy2019fully,
  title={Fully convolutional geometric features},
  author={Choy, Christopher and Park, Jaesik and Koltun, Vladlen},
  booktitle={Proceedings of the IEEE/CVF international conference on computer vision},
  pages={8958--8966},
  year={2019}
}

@article{2D3Dproj_ICCV2025_sam4d,
  title={SAM4D: Segment Anything in Camera and LiDAR Streams},
  author={Xu, Jianyun and Wang, Song and Ni, Ziqian and Hu, Chunyong and Yang, Sheng and Zhu, Jianke and Li, Qiang},
  journal={arXiv preprint arXiv:2506.21547},
  year={2025}
}

@inproceedings{2D3Dproj_partslip,
  title={Partslip: Low-shot part segmentation for 3d point clouds via pretrained image-language models},
  author={Liu, Minghua and Zhu, Yinhao and Cai, Hong and Han, Shizhong and Ling, Zhan and Porikli, Fatih and Su, Hao},
  booktitle={Proceedings of the IEEE/CVF conference on computer vision and pattern recognition},
  pages={21736--21746},
  year={2023}
}

@article{2D3Dproj_partslip++,
  title={Partslip++: Enhancing low-shot 3d part segmentation via multi-view instance segmentation and maximum likelihood estimation},
  author={Zhou, Yuchen and Gu, Jiayuan and Li, Xuanlin and Liu, Minghua and Fang, Yunhao and Su, Hao},
  journal={arXiv preprint arXiv:2312.03015},
  year={2023}
}

@article{2D3Dproj_ICCVw2023_sam3d,
  title={Sam3d: Segment anything in 3d scenes},
  author={Yang, Yunhan and Wu, Xiaoyang and He, Tong and Zhao, Hengshuang and Liu, Xihui},
  journal={arXiv preprint arXiv:2306.03908},
  year={2023}
}

@article{2D3Dproj_neurips2025_sa3dip,
  title={SA3DIP: Segment Any 3D Instance with Potential 3D Priors},
  author={Yang, Xi and Gu, Xu and Yin, Xingyilang and Gao, Xinbo},
  journal={Advances in Neural Information Processing Systems},
  volume={37},
  pages={102568--102586},
  year={2024}
}

@article{2D3Dproj_IROS2025_labelefficientLPS,
  title={Label-Efficient LiDAR Panoptic Segmentation},
  author={{\c{C}}anak{\c{c}}{\i}, Ahmet Selim and V{\"o}disch, Niclas and Petek, K{\"u}rsat and Burgard, Wolfram and Valada, Abhinav},
  journal={arXiv preprint arXiv:2503.02372},
  year={2025}
}

@inproceedings{2D3Dproj_autoprosam,
  title={Autoprosam: Automated prompting sam for 3d multi-organ segmentation},
  author={Li, Chengyin and Sultan, Rafi Ibn and Khanduri, Prashant and Qiang, Yao and Indrin, Chetty and Zhu, Dongxiao},
  booktitle={2025 IEEE/CVF Winter Conference on Applications of Computer Vision (WACV)},
  pages={3570--3580},
  year={2025},
  organization={IEEE}
}

@article{2D3Dproj_SA3D,
  title={Segment anything in 3d with nerfs},
  author={Cen, Jiazhong and Zhou, Zanwei and Fang, Jiemin and Shen, Wei and Xie, Lingxi and Jiang, Dongsheng and Zhang, Xiaopeng and Tian, Qi and others},
  journal={Advances in Neural Information Processing Systems},
  volume={36},
  pages={25971--25990},
  year={2023}
}

@inproceedings{2D3Dproj_sai3d,
  title={Sai3d: Segment any instance in 3d scenes},
  author={Yin, Yingda and Liu, Yuzheng and Xiao, Yang and Cohen-Or, Daniel and Huang, Jingwei and Chen, Baoquan},
  booktitle={Proceedings of the IEEE/CVF Conference on Computer Vision and Pattern Recognition},
  pages={3292--3302},
  year={2024}
}

@inproceedings{2D3Dproj_pointseg,
  title={Pointseg: A training-free paradigm for 3d scene segmentation via foundation models},
  author={He, Qingdong and Peng, Jinlong and Jiang, Zhengkai and Hu, Xiaobin and Zhang, Jiangning},
  booktitle={Proceedings of the IEEE/CVF International Conference on Computer Vision},
  pages={2657--2667},
  year={2025}
}

@inproceedings{2D3Dproj_partfield,
  title={Partfield: Learning 3d feature fields for part segmentation and beyond},
  author={Liu, Minghua and Uy, Mikaela Angelina and Xiang, Donglai and Su, Hao and Fidler, Sanja and Sharp, Nicholas and Gao, Jun},
  booktitle={Proceedings of the IEEE/CVF International Conference on Computer Vision},
  pages={9704--9715},
  year={2025}
}

@article{2D3Dproj_sampart3d,
  title={Sampart3d: Segment any part in 3d objects},
  author={Yang, Yunhan and Huang, Yukun and Guo, Yuan-Chen and Lu, Liangjun and Wu, Xiaoyang and Lam, Edmund Y and Cao, Yan-Pei and Liu, Xihui},
  journal={arXiv preprint arXiv:2411.07184},
  year={2024}
}

@inproceedings{li2025ga,
  title={GA-SAM: Geometry-Aware SAM Adaptation with Sparse Annotation-Driven Point Cloud Completion},
  author={Li, Shumeng and Zhang, Jian and Qi, Lei and Shi, Yinghuan},
  booktitle={International Conference on Medical Image Computing and Computer-Assisted Intervention},
  pages={214--224},
  year={2025},
  organization={Springer}
}

@inproceedings{thomas2019kpconv,
  title={Kpconv: Flexible and deformable convolution for point clouds},
  author={Thomas, Hugues and Qi, Charles R and Deschaud, Jean-Emmanuel and Marcotegui, Beatriz and Goulette, Fran{\c{c}}ois and Guibas, Leonidas J},
  booktitle={Proceedings of the IEEE/CVF international conference on computer vision},
  pages={6411--6420},
  year={2019}
}

@inproceedings{zhu2021cylindrical,
  title={Cylindrical and asymmetrical 3d convolution networks for lidar segmentation},
  author={Zhu, Xinge and Zhou, Hui and Wang, Tai and Hong, Fangzhou and Ma, Yuexin and Li, Wei and Li, Hongsheng and Lin, Dahua},
  booktitle={Proceedings of the IEEE/CVF conference on computer vision and pattern recognition},
  pages={9939--9948},
  year={2021}
}

@article{zhou2020cylinder3d,
  title={Cylinder3d: An effective 3d framework for driving-scene lidar semantic segmentation},
  author={Zhou, Hui and Zhu, Xinge and Song, Xiao and Ma, Yuexin and Wang, Zhe and Li, Hongsheng and Lin, Dahua},
  journal={arXiv preprint arXiv:2008.01550},
  year={2020}
}

@article{marcuzzi2023mask4d,
  title={Mask4d: End-to-end mask-based 4d panoptic segmentation for lidar sequences},
  author={Marcuzzi, Rodrigo and Nunes, Lucas and Wiesmann, Louis and Marks, Elias and Behley, Jens and Stachniss, Cyrill},
  journal={IEEE Robotics and Automation Letters},
  volume={8},
  number={11},
  pages={7487--7494},
  year={2023},
  publisher={IEEE}
}

@inproceedings{kreuzberg20224d,
  title={4d-stop: Panoptic segmentation of 4d lidar using spatio-temporal object proposal generation and aggregation},
  author={Kreuzberg, Lars and Zulfikar, Idil Esen and Mahadevan, Sabarinath and Engelmann, Francis and Leibe, Bastian},
  booktitle={European Conference on Computer Vision},
  pages={537--553},
  year={2022},
  organization={Springer}
}

@article{mersch2022receding,
  title={Receding moving object segmentation in 3d lidar data using sparse 4d convolutions},
  author={Mersch, Benedikt and Chen, Xieyuanli and Vizzo, Ignacio and Nunes, Lucas and Behley, Jens and Stachniss, Cyrill},
  journal={IEEE Robotics and Automation Letters},
  volume={7},
  number={3},
  pages={7503--7510},
  year={2022},
  publisher={IEEE}
}

@article{wang2025segnet4d,
  title={SegNet4D: Efficient Instance-Aware 4D Semantic Segmentation for LiDAR Point Cloud},
  author={Wang, Neng and Guo, Ruibin and Shi, Chenghao and Wang, Ziyue and Zhang, Hui and Lu, Huimin and Zheng, Zhiqiang and Chen, Xieyuanli},
  journal={IEEE Transactions on Automation Science and Engineering},
  year={2025},
  publisher={IEEE}
}

@inproceedings{zhu20234d_eq4dpls,
  title={4d panoptic segmentation as invariant and equivariant field prediction},
  author={Zhu, Minghan and Han, Shizhong and Cai, Hong and Borse, Shubhankar and Ghaffari, Maani and Porikli, Fatih},
  booktitle={Proceedings of the IEEE/CVF International Conference on Computer Vision},
  pages={22488--22498},
  year={2023}
}

@article{marcuzzi2022_CA_net,
  title={Contrastive instance association for 4d panoptic segmentation using sequences of 3d lidar scans},
  author={Marcuzzi, Rodrigo and Nunes, Lucas and Wiesmann, Louis and Vizzo, Ignacio and Behley, Jens and Stachniss, Cyrill},
  journal={IEEE Robotics and Automation Letters},
  volume={7},
  number={2},
  pages={1550--1557},
  year={2022},
  publisher={IEEE}
}

@article{hu2022lora,
  title={Lora: Low-rank adaptation of large language models.},
  author={Hu, Edward J and Shen, Yelong and Wallis, Phillip and Allen-Zhu, Zeyuan and Li, Yuanzhi and Wang, Shean and Wang, Lu and Chen, Weizhu and others},
  journal={ICLR},
  volume={1},
  number={2},
  pages={3},
  year={2022}
}

@inproceedings{li2023memoryseg,
  title={Memoryseg: Online lidar semantic segmentation with a latent memory},
  author={Li, Enxu and Casas, Sergio and Urtasun, Raquel},
  booktitle={Proceedings of the IEEE/CVF International Conference on Computer Vision},
  pages={745--754},
  year={2023}
}

@inproceedings{liu2023mars3d,
  title={Mars3d: A plug-and-play motion-aware model for semantic segmentation on multi-scan 3d point clouds},
  author={Liu, Jiahui and Chang, Chirui and Liu, Jianhui and Wu, Xiaoyang and Ma, Lan and Qi, Xiaojuan},
  booktitle={Proceedings of the IEEE/CVF Conference on Computer Vision and Pattern Recognition},
  pages={9372--9381},
  year={2023}
}

@article{shi2024learning,
  title={Learning temporal variations for 4D point cloud segmentation},
  author={Shi, Hanyu and Wei, Jiacheng and Wang, Hao and Liu, Fayao and Lin, Guosheng},
  journal={International Journal of Computer Vision},
  volume={132},
  number={12},
  pages={5603--5617},
  year={2024},
  publisher={Springer}
}

@inproceedings{shi2020spsequencenet,
  title={Spsequencenet: Semantic segmentation network on 4d point clouds},
  author={Shi, Hanyu and Lin, Guosheng and Wang, Hao and Hung, Tzu-Yi and Wang, Zhenhua},
  booktitle={Proceedings of the IEEE/CVF conference on computer vision and pattern recognition},
  pages={4574--4583},
  year={2020}
}

@inproceedings{duerr2020lidar,
  title={Lidar-based recurrent 3d semantic segmentation with temporal memory alignment},
  author={Duerr, Fabian and Pfaller, Mario and Weigel, Hendrik and Beyerer, J{\"u}rgen},
  booktitle={2020 International Conference on 3D Vision (3DV)},
  pages={781--790},
  year={2020},
  organization={IEEE}
}

@inproceedings{wu2024taseg,
  title={Taseg: Temporal aggregation network for lidar semantic segmentation},
  author={Wu, Xiaopei and Hou, Yuenan and Huang, Xiaoshui and Lin, Binbin and He, Tong and Zhu, Xinge and Ma, Yuexin and Wu, Boxi and Liu, Haifeng and Cai, Deng and others},
  booktitle={Proceedings of the IEEE/CVF Conference on Computer Vision and Pattern Recognition},
  pages={15311--15320},
  year={2024}
}

@inproceedings{wang2023insmos,
  title={Insmos: Instance-aware moving object segmentation in lidar data},
  author={Wang, Neng and Shi, Chenghao and Guo, Ruibin and Lu, Huimin and Zheng, Zhiqiang and Chen, Xieyuanli},
  booktitle={2023 IEEE/RSJ International Conference on Intelligent Robots and Systems (IROS)},
  pages={7598--7605},
  year={2023},
  organization={IEEE}
}

@article{llava,
  title={Visual instruction tuning},
  author={Liu, Haotian and Li, Chunyuan and Wu, Qingyang and Lee, Yong Jae},
  journal={Advances in neural information processing systems},
  volume={36},
  pages={34892--34916},
  year={2023}
}

@inproceedings{blip2,
  title={Blip-2: Bootstrapping language-image pre-training with frozen image encoders and large language models},
  author={Li, Junnan and Li, Dongxu and Savarese, Silvio and Hoi, Steven},
  booktitle={International conference on machine learning},
  pages={19730--19742},
  year={2023},
  organization={PMLR}
}

@article{chen2021moving,
  title={Moving object segmentation in 3D LiDAR data: A learning-based approach exploiting sequential data},
  author={Chen, Xieyuanli and Li, Shijie and Mersch, Benedikt and Wiesmann, Louis and Gall, J{\"u}rgen and Behley, Jens and Stachniss, Cyrill},
  journal={IEEE Robotics and Automation Letters},
  volume={6},
  number={4},
  pages={6529--6536},
  year={2021},
  publisher={IEEE}
}

@article{rozsa2025efficient,
  title={Efficient moving object segmentation in LiDAR point clouds using minimal number of sweeps},
  author={Rozsa, Zoltan and Madaras, Akos and Sziranyi, Tamas},
  journal={IEEE Open Journal of Signal Processing},
  year={2025},
  publisher={IEEE}
}

@inproceedings{zeng2024mambamos,
  title={Mambamos: Lidar-based 3d moving object segmentation with motion-aware state space model},
  author={Zeng, Kang and Shi, Hao and Lin, Jiacheng and Li, Siyu and Cheng, Jintao and Wang, Kaiwei and Li, Zhiyong and Yang, Kailun},
  booktitle={Proceedings of the 32nd ACM International Conference on Multimedia},
  pages={1505--1513},
  year={2024}
}

@inproceedings{kweon2024weakly,
  title={Weakly supervised point cloud semantic segmentation via artificial oracle},
  author={Kweon, Hyeokjun and Kim, Jihun and Yoon, Kuk-Jin},
  booktitle={Proceedings of the IEEE/CVF Conference on Computer Vision and Pattern Recognition},
  pages={3721--3731},
  year={2024}
}
\end{document}